\documentclass[final]{cvpr}

\usepackage{times}
\usepackage{epsfig}
\usepackage{graphicx}
\usepackage{amsmath}
\usepackage{amssymb}

\usepackage{amsfonts}
\usepackage{tabularx}
\usepackage{amsmath}
\usepackage{booktabs}
\usepackage{multirow}
\usepackage{xcolor}
\usepackage[switch]{lineno}
\usepackage{subcaption}
\usepackage{bm}

\usepackage[pagebackref=true,breaklinks=true,colorlinks,bookmarks=false]{hyperref}

\def\cvprPaperID{375} 
\def\confYear{CVPR 2021}

\begin{document}


\title{Wide-Baseline Multi-Camera Calibration using Person Re-Identification}

\author{Yan Xu\qquad\qquad Yu-Jhe Li\qquad\qquad Xinshuo Weng\qquad\qquad Kris Kitani\\
Carnegie Mellon University\\
{\tt\small \{yxu2, yujheli, xinshuow, kkitani\}@cs.cmu.edu}
}

\maketitle

\begin{abstract}
We address the problem of estimating the 3D pose of a network of cameras for large-environment wide-baseline scenarios, e.g., cameras for construction sites, sports stadiums, and public spaces. This task is challenging since detecting and matching the same 3D keypoint observed from two very different camera views is difficult, making standard structure-from-motion (SfM) pipelines inapplicable. In such circumstances, treating people in the scene as ``keypoints" and associating them across different camera views can be an alternative method for obtaining correspondences. Based on this intuition, we propose a method that uses ideas from person re-identification (re-ID) for wide-baseline camera calibration. Our method first employs a re-ID method to associate human bounding boxes across cameras, then converts bounding box correspondences to point correspondences, and finally solves for camera pose using multi-view geometry and bundle adjustment. Since our method does not require specialized calibration targets except for visible people, it applies to situations where frequent calibration updates are required. We perform extensive experiments on datasets captured from scenes of different sizes ($80m^2, 350m^2, 600m^2$), camera settings (indoor and outdoor), and human activities (walking, playing basketball, construction). Experiment results show that our method achieves similar performance to standard SfM methods relying on manually labeled point correspondences.
\end{abstract}

\section{Introduction}
\begin{figure}[t]
\centering
\includegraphics[width=0.95\linewidth]{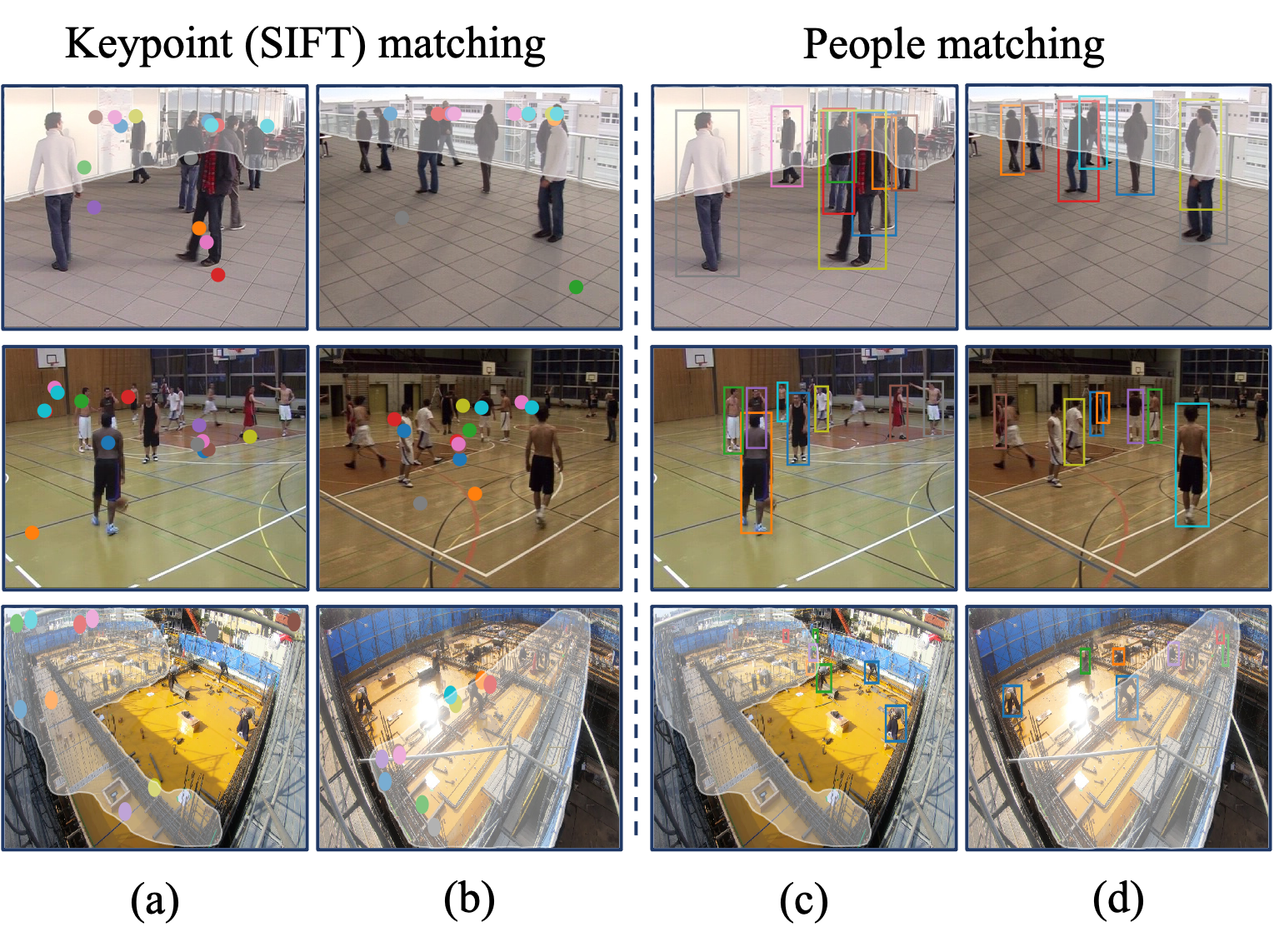} 
\vspace{-0.2cm}
\caption{In large-environment, wide-baseline camera networks, backgrounds (top, white area), lighting conditions (middle), and textured areas (bottom, white area) from different camera views can vary massively, causing the failure of keypoint detection and matching ((a), (b)) in standard SfM methods.  However, matching the same people across views can still be done correctly using re-ID methods ((c), (d)). (Points/bounding boxes of the same color correspond.)\vspace{-5pt}}
\label{fig:teaser}
\end{figure}
\begin{figure*}[htbp]
    \centering
    \begin{subfigure}[b]{0.28\textwidth}
        \includegraphics[width=0.95\textwidth]{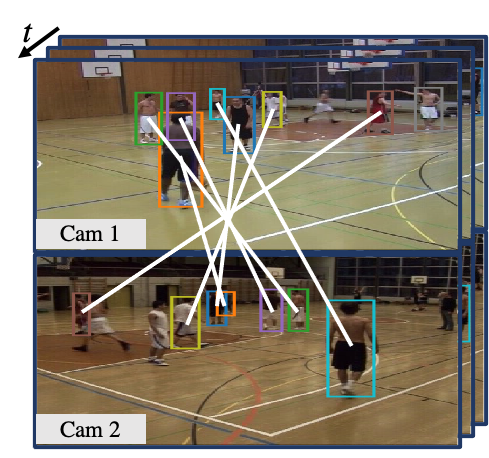}
        \caption{Bounding box correspondences}
        \label{subfig:bbox_corresp}
    \end{subfigure}
    \hskip 2ex
    \begin{subfigure}[b]{0.28\textwidth}
        \includegraphics[width=0.95\textwidth]{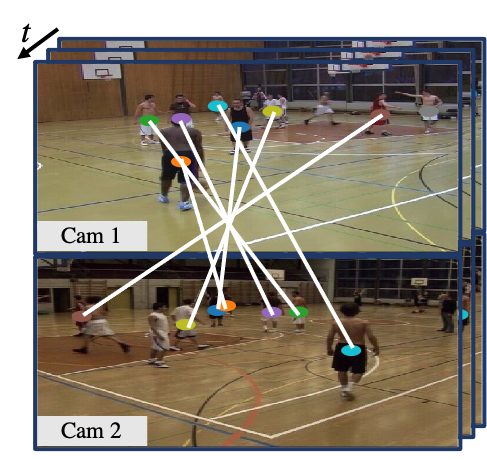}
        \caption{Point correspondences}
        \label{subfig:pts_corresp}
    \end{subfigure}
    \hskip 2ex
    \begin{subfigure}[b]{0.28\textwidth}
        \includegraphics[width=0.95\textwidth]{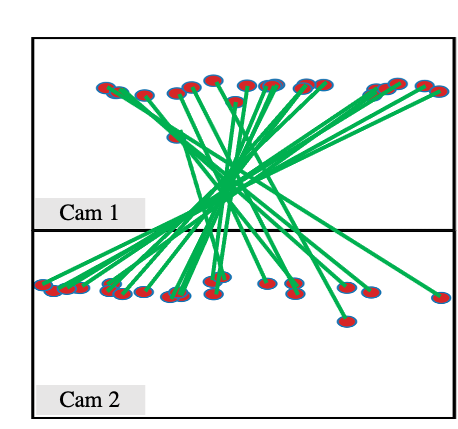}
        \caption{Point correspondences over time}
        \label{subfig:pts_corresp_all}
    \end{subfigure}
    \caption[]{\textbf{Method overview.} Given a set of synchronized videos from different views and detected human bounding boxes, our method first associates bounding boxes across camera views using a re-ID network (a), then converts the bounding box correspondences to 2D-2D point correspondences (b).  Finally, our method aggregates the point correspondences over time (c) and solves the camera pose with the point correspondences as input using multi-view geometry and bundle adjustment.}
    \label{fig:correspondence}
\end{figure*}
Our task is to solve the 3D camera pose estimation problem for multi-camera networks. We target challenging large-environment wide-baseline scenarios where cameras are static, sparse, and spaced far from each other (\eg, 10 to 20 meters).
Conventionally, SfM~\cite{ullman1979interpretation} methods are often used to estimate camera pose.
These methods first detect keypoints in the images of different camera views and describe the keypoints using hand-crafted or deep features~\cite{lowe2004distinctive, detone2018superpoint}.
They then match the keypoint features~\cite{jakubovic2018image} across views to obtain 2D-2D  point correspondences and solve the camera pose using multi-view geometry~\cite{hartley2003multiple}.
In our setting, due to the significant difference between camera poses, images taken from different cameras can have very different backgrounds, lighting conditions, and texture areas, as shown in Figure \ref{fig:teaser}, making detecting and matching the same keypoints across camera views difficult.

Alternatively, we observe that if people are visible in the scene, we can obtain correspondences by detecting and matching people across different cameras. Especially for wide-baseline scenarios, matching the same people across camera views is easier than matching 2D keypoints since many human features can be used. For example, when people are close to the cameras, we can use the appearance information (\eg, height, clothes, length of hair) to match them across camera views (Figure \ref{fig:teaser}). When people are far away from the cameras, we can still successfully match them from their temporal motion information (\eg, speed, smoothness of trajectory). Therefore, to obtain correspondences in wide-baseline scenarios, we can treat people as ``keypoints" and associate their bounding boxes across camera views using a re-ID algorithm. Nevertheless, simply obtaining the associated bounding boxes is not enough to establish accurate 2D-2D correspondences necessary for camera pose estimation. As a second step, we can associate the same body parts (\eg, head, foot, the position of body mass) inside the bounding boxes to further obtain point correspondences. In short, we can solve the feature matching problem by breaking it into a two-step process: First, matching people across camera views; Second, converting the bounding box correspondences to point correspondences. Since we assume that people are moving in the scene, which is often true, we can obtain a sufficient number of point correspondences for solving camera poses by aggregating the correspondences from a sequence of video frames.  

We summarize our proposed wide-baseline camera pose estimation method in Figure \ref{fig:correspondence}.  Our method includes three stages: 1) person matching, 2) point correspondence extraction, and 3) geometric camera pose estimation. Given a set of synchronized videos captured by cameras of different views, whose intrinsic and distortion parameters are given from the previous calibration, our method first associates person bounding boxes using a re-ID network pre-trained on open datasets~\cite{solera2016tracking, zheng2015scalable}. As a second step, our method converts the bounding box correspondences to point correspondences by extracting and associating the bounding box centers, which approximate the body mass positions. Finally, our method aggregates the point correspondences over time and solves for camera poses using a structure-from-motion pipeline (algebraic estimates of pose pairwise cameras, followed by non-linear optimization via bundle adjustment).

Our method only assumes the existence of visible moving people in the scene without the requirement of any other specialized calibration targets. It is thus suitable for many situations where consistent camera pose estimation is required, \eg, basketball training where cameras need to be moved for each new game or construction sites where cameras must be moved as the site is constructed. Moreover, our method does not require the re-ID model to output perfect association results since  RANSAC~\cite{fischler1981random} in the later stage can filter outliers. We evaluate our method on three datasets collected from scenes of different sizes ($80m^2, 350m^2, 600m^2$) and lighting conditions (indoor and outdoor). The human postures in the three datasets also vary significantly, including walking, shooting, running, jumping, crouching, \etc. We aim to use these datasets of various environment settings and human postures to evaluate the robustness of our method. Experiments show that our method achieves similar performance to a standard SfM pipeline which relies on manually labeled point correspondences.

Our contributions are as follows: 1) We propose to apply person re-ID algorithms to solve camera pose estimation for multi-camera networks. 2) We contribute a two-step process treating people as ``keypoints" to obtain correspondences for wide-baseline scenarios. 3) Our method achieves an average accuracy of $(0.4m, 1.08^{\circ})$ across three datasets, comparable with SfM methods using manual annotation. 4) We perform extensive robustness and efficiency analysis for a more comprehensive understanding of our method.

\section{Related work}
\label{sec:related_work}

Methods for solving the relative camera pose estimation problem can generally be categorized into geometric methods using the SfM pipeline and end-to-end deep pose regressors. We will discuss both categories of methods and briefly review recent deep re-ID works in this section.

\vspace{2mm}\noindent\textbf{Geometric methods} address the camera pose estimation problem with a two-stage framework by first obtaining 2D-2D  point correspondences and then solving the camera pose using a geometric pipeline.
Generally, they first detect the keypoints (Harris~\cite{harris1988combined}, FAST~\cite{rosten2006machine}, \etc) in the images of different views and describe the keypoints with hand-crafted features (SIFT~\cite{lowe2004distinctive}, BRIEF~\cite{calonder2010brief}, ORB~\cite{rublee2011orb}, \etc). Then, they match the keypoints (BFM~\cite{jakubovic2018image}, FLANN \cite{muja2009flann}, \etc) across images to obtain point correspondences. 
Recently, many deep learning methods solve the keypoints detection and description simultaneously using neural networks (SuperPoint~\cite{detone2018superpoint}, UR2KiD~\cite{yang2020ur2kid}, D2-net~\cite{dusmanu2019d2}, LIFT~\cite{yi2016lift}, Lf-net~\cite{ono2018lf}, Elf~\cite{benbihi2019elf}). 
Once obtaining 2D-2D point correspondences, these methods solve camera poses using a multi-view geometry pipeline.
They first use the N-point algorithm~\cite{hartley2003multiple, li2006five, nister2004efficient}, usually inside a RANSAC~\cite{fischler1981random} loop, to solve for the Essential matrix, which will then be decomposed to the camera rotation and an up-to-scale camera translation~\cite{georgiev2014practical}.
As a final step, they use bundle adjustment~\cite{triggs1999bundle} to further optimize the 3D poses of all cameras.
There are also works~\cite{bao2011semantic, dame2013dense} using semantic information for solving SfM problems.
These geometric methods are quite mature and generally accurate.
However, they have difficulty in matching keypoint features across camera views when the distance between cameras is vast, as in our task.
In this work, we use person matching to obtain correspondences to solve camera poses for wide-baseline scenarios.

\vspace{2mm}\noindent\textbf{Deep pose regressor} was first applied in absolute camera pose estimation.
PoseNet \cite{kendall2015posenet} is the first attempt that treats camera pose estimation as an end-to-end regression problem and solves the problem with a convolutional neural network~\cite{gu2018recent}, trained on data labeled using SfM~\cite{ullman1979interpretation}. Since then, many improvements on PoseNet have been proposed, including using different network architectures \cite{wu2017delving, naseer2017deep, melekhov2017image} and new loss designs \cite{kendall2017geometric, laskar2017camera, brahmbhatt2018geometry}.
Deep pose regressors for relative camera pose estimation~\cite{melekhov2017relative, laskar2017camera, balntas2018relocnet, ding2019camnet} have also been applied recently. In General, these methods input a pair of images into a Siamese network architecture\cite{melekhov2017relative} to extract deep features, from which they regress the camera pose.
Despite the convenience of end-to-end regression, deep regressors  still maintain a performance gap compared to geometric methods~\cite{shavit2019introduction}.
Moreover, deep pose regressors require images taken from moving cameras for training.  In our task, the cameras are static, meaning that the training images and test images would be almost the same, making the deep pose regressors overfit to one camera pose. Therefore, deep pose regressors are inapplicable in our task.

\vspace{2mm}\noindent\textbf{Deep re-identification} has been used for people matching in our method.
Many existing deep re-ID algorithms, e.g.,\ \cite{lin2017improving,shen2018deep,shen2018person,kalayeh2018human,cheng2016person,chang2018multi,chen2018group,sun2018beyond,suh2018part}, are developed to address various challenges in re-ID problem, such as background clutter, viewpoint changes, and pose variations. For instance, Yang~\etal~\cite{zhong2017camera} learn a camera-invariant subspace to deal with the style variations caused by different cameras. Liu~\etal~\cite{liu2018pose} develop a pose-transferable framework based on generative adversarial network (GAN)~\cite{goodfellow2014generative} to yield pose-specific images for tackling pose variations. Several methods addressing background clutter leverage attention mechanisms to emphasize the discriminative parts~\cite{li2018harmonious,song2018mask,si2018dual}. In addition to these methods that learn global features, a few methods further utilize part-level information~\cite{sun2018beyond} to learn more fine-grained features, adopt human semantic parsing for learning local features~\cite{kalayeh2018human}, or derive part-aligned representations~\cite{suh2018part} for improving person re-ID.
Following these works, we choose the most commonly used model~\cite{zheng2016person} with ResNet-50~\cite{he2016deep} as the backbone of our re-ID network.

\section{Method}
\begin{figure*}
  \centering
  \includegraphics[width=0.93\linewidth]{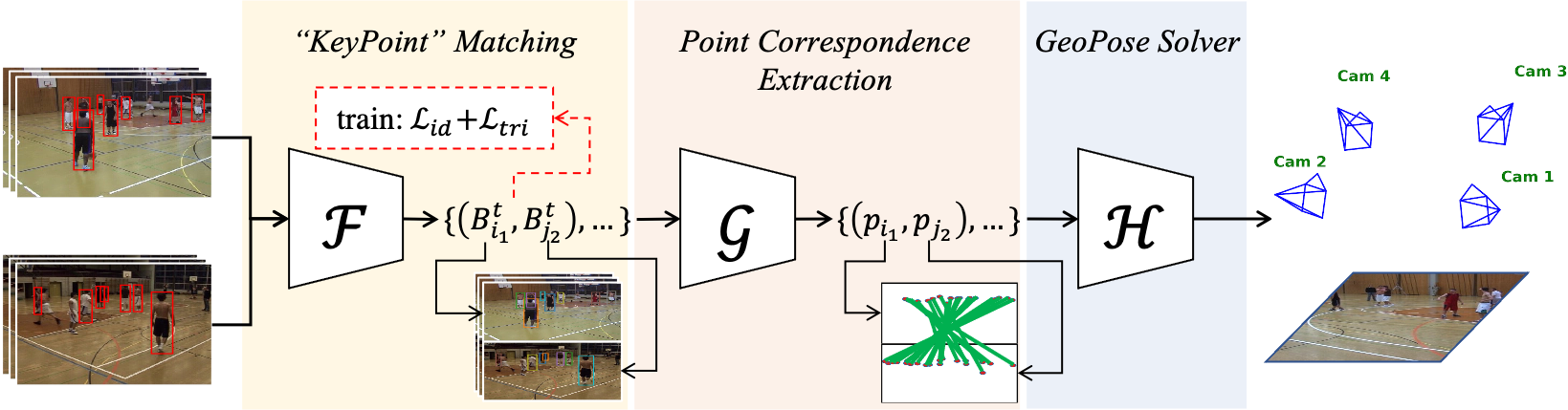}
  \caption{\textbf{System pipeline.}  Our method includes three stages.  First, we use a re-ID network ($\mathcal{F}$) to associate person bounding boxes across camera views. Second, we use a function ($\mathcal{G}$) to convert bounding box correspondences to point correspondences.  Finally, we solve the camera pose using a GeoPose solver ($\mathcal{H}$).  $B^t_{i_1}$ is the $i$-th bounding box of camera 1 at frame $t$, $p_{i_1}$ is the $i$-th keypoint of camera 1, the definitions of $B^t_{j_2}$ and $p_{j_2}$ are similar to $B^t_{i_1}$ and $p_{i_1}$, respectively.\vspace{-5pt}}
  \label{fig:system_pipe}
\end{figure*}

To solve the camera pose estimation problem for large-environment, wide-baseline scenarios, we propose to treat people as ``keypoints" and use person re-ID for obtaining 2D-2D correspondences. Figure \ref{fig:system_pipe} presents the system pipeline of our method, which includes three modules: 1) ``keypoint" matching, 2) point correspondence extraction, and 3) the geometric camera pose (GeoPose) solver.

\subsection{``Keypoint" Matching Using Person Re-ID}
The first module of our method is ``keypoint" matching.  We adopt the re-ID model in the previous work~\cite{zheng2016person} using ResNet-50~\cite{he2016deep} as the backbone.
As shown in Figure~\ref{fig:system_pipe}, the learning of the re-ID model is guided by a person id classification loss $\mathcal{L}_{id}$ and a discriminative triplet loss $\mathcal{L}_{tri}$.

\noindent\textbf{Training: } At the training stage, we have an image set $X = \{x_i\}_{i=1}^N$ and its corresponding label set $Y = \{y_i\}_{i=1}^N$ with size~$N$, where $x_i \in {R}^{H \times W \times 3}$ and $y_i \in \mathbb{R}$.
We first employ the classification loss $\mathcal{L}_{id}$ by computing the negative log-likelihood between the predicted label $\tilde{y} \in \mathbb{R}^K$ and the ground truth $\hat{y} \in \mathbb{N}^K$:
\begin{equation}
  \mathcal{L}_{id} =  - \mathbb{E}_{(x,y) \sim (X,Y)}\sum_{k=1}^{K}\hat{y}_k\log(\tilde{y}_k)\\
  \label{eq:cls}
\end{equation}
where $K$ is the number of identities (classes).

To further enhance the discriminative property, we impose a triplet loss $\mathcal{L}_{tri}$, which maximizes the inter-class discrepancy while minimizing intra-class distinctness. Specifically, for each input image $x$, we sample a positive image $x_\mathrm{pos}$ with the same identity label and a negative image $x_\mathrm{neg}$ with different identity labels to form a triplet tuple. The distances between $x$ and $x_\mathrm{pos}$/$x_\mathrm{neg}$ can be computed as
\begin{align}
    d_\mathrm{pos} = \|{f}_x - {f}_{x_\mathrm{pos}}\|_2\\
    d_\mathrm{neg} = \|{f}_x - {f}_{x_\mathrm{neg}}\|_2
\end{align}
where ${f}_x$, ${f}_{x_\mathrm{pos}}$, and ${f}_{x_\mathrm{neg}}$ represent the feature vectors of images $x$, $x_\mathrm{pos}$, and $x_\mathrm{neg}$, respectively. We then can have the triplet loss $\mathcal{L}_{tri}$ defined as
\begin{equation}
  \mathcal{L}_{tri} = \mathbb{E}_{(x,y) \sim (X,Y)}\max(0, m + d_\mathrm{pos} - d_\mathrm{neg}) \\
  \label{eq:tri}
\end{equation}
where $m > 0$ is the margin used to define the difference between the distance of positive image pair $d_\mathrm{pos}$ and the distance of negative image pair $d_\mathrm{neg}$.

In this work, we allow, to a certain extent, the imperfectness of the re-ID model and use RANSAC in the GeoPose solver to reject the mis-associations.
We pre-train the re-ID model on open datasets~\cite{solera2016tracking, zheng2015scalable}.
For the relatively easier datasets, \textit{\textit{Terrace}} and \textit{
\textit{Basketball}}~\cite{fleuret2007multicamera}, we directly apply the pre-trained model for inference.
We also collected a more challenging dataset, \textit{\textit{ConstructSite}}, on which we fine-tune the pre-trained re-ID model first before inference.

\noindent\textbf{Inference: } At the inference stage, we use the re-ID model to associate bounding boxes across cameras. In this work, we assume that person trajectories in each video are provided.  To extract re-ID feature of a person bounding box $b_i$ in frame $i$, we first apply the restricted random sampling strategy~\cite{li2018diversity} on the whole tracklet $\mathcal{T}$ (a sequence of bounding boxes) of this person to obtain a smaller tracklet $\mathcal{T}_i = \{b_k\}_{k=1}^{8}$. We then extract the re-ID feature from bounding boxes in $\mathcal{T}_i$ and do an average-pooling to get the feature representation of $b_i$.  Finally, we apply the Hungarian algorithm~\cite{kuhn1955hungarian} to match bounding boxes across cameras.

\subsection{Point Correspondence Extraction}
After obtaining bounding box correspondences, our next step is to find 2D-2D point correspondences from the bounding box correspondences, such that the matched point correspondences are semantically meaningful.  The intuitive idea is to associate the same body part (\eg head, foot, the center of body mass) inside the bounding boxes.  Our idea leverages the critical observation that the bounding box center can serve as a rough estimate of the body mass position.

\begin{figure}[t]
\centering
\includegraphics[width=0.95\columnwidth]{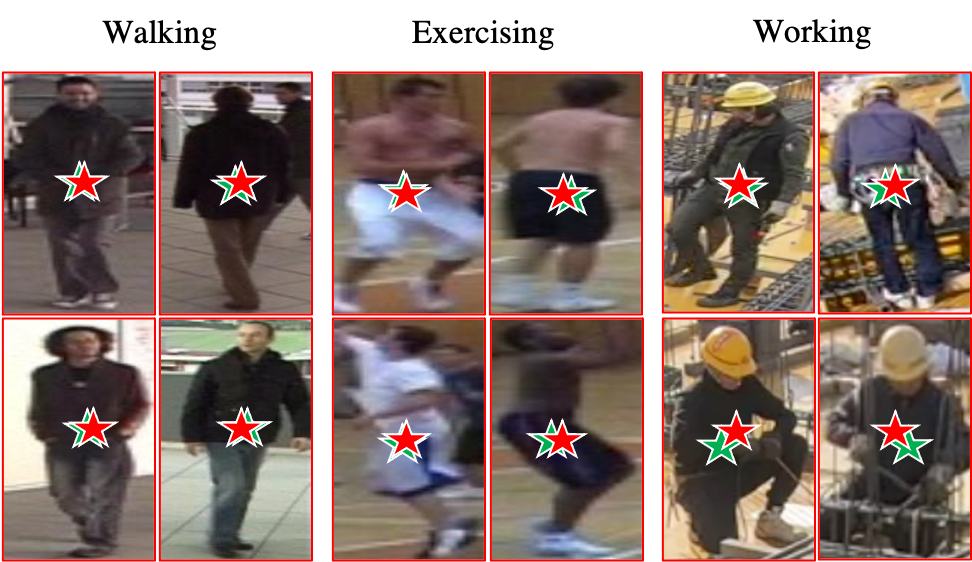}
\caption{\textbf{Body mass positions (green) and bounding box centers (red).}  We present samples from three datasets of variant camera poses and human postures and observe that bounding box centers can approximate body mass positions.\vspace{-5pt}}
\label{fig:bbox_point_corresp}
\end{figure}

We visualize in Figure \ref{fig:bbox_point_corresp} the bounding box centers and the centers of body mass of various human poses.  The visualized person poses include people walking, sporting (running, jumping, shooting a basketball \etc), and working (bending, carrying, crouching, \etc), captured from different camera heights and view angles.  As Figure \ref{fig:bbox_point_corresp} shows, the bounding box center is close to the center of body mass under different camera poses and human postures.  We thus use the bounding box center to represent the center of body mass and associate them across camera views as 2D-2D  point correspondences.  However, such an approximation would not be perfect. There will be an offset between the position of the bounding box center and the center of body mass, meaning that the obtained point correspondences will be noisy.  Our method uses RANSAC~\cite{fischler1981random} to reject the correspondences with large position offsets.  For other point correspondences, our method treats their position offsets as noise and optimizes over all the point correspondences using bundle adjustment~\cite{triggs1999bundle} to minimize the impact.

Formally, we define a bounding box as $[u_{tl}, v_{tl}, u_{br}, v_{br}]$, in which $[u_{tl}, v_{tl}]$ and $[u_{br}, v_{br}]$ represent the top-left and bottom-right corners of the bounding box.  We can then obtain the position of the body mass center $[u, v]$ as :
\begin{equation}
    [u, v] = [(u_{tl} + v_{br}) / 2, (u_{tl} + v_{br})/2]\\
\end{equation}

\subsection{Geometric Camera Pose (GeoPose) Solver}
\begin{figure}[t]
\centering
\includegraphics[width=0.95\columnwidth]{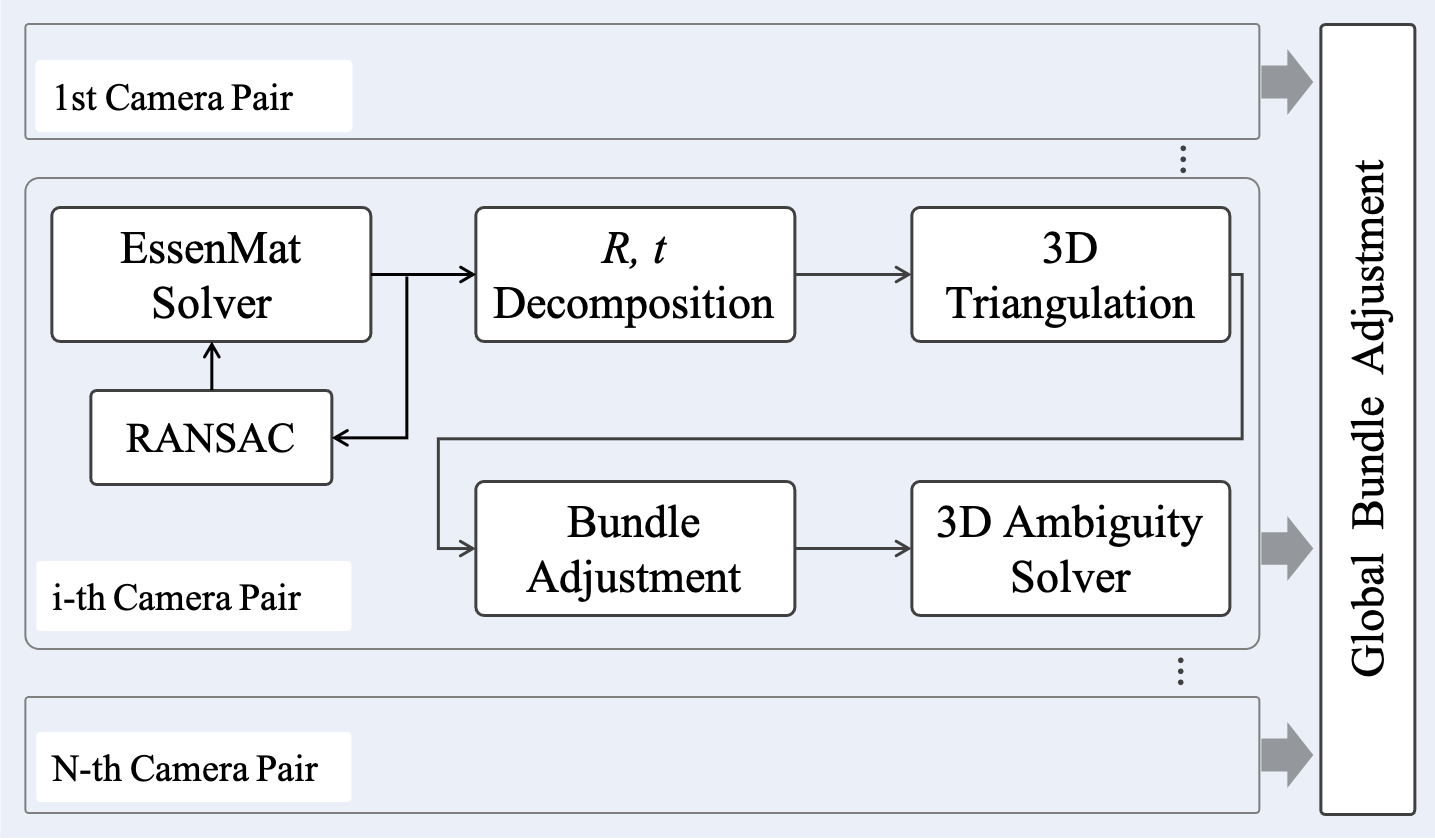}
\caption{\textbf{GeoPose solver.}  We first solve the relative camera pose for each camera pair using a five-step process, then optimize all camera poses with a global bundle adjustment.\vspace{-5pt}}
\label{fig:geo_pipeline}
\end{figure}

With the 2D-2D point correspondences, our final step is to solve the camera pose with our GeoPose solver.
Figure \ref{fig:geo_pipeline} shows the diagram of the GeoPose solver.
We first solve the relative pose for each camera pair with a five-step pipeline, then optimize all camera poses using a global bundle adjustment.
Note that we assume the camera intrinsic and distortion parameters are provided.
Moreover, traditional automatic feature matching for resectioning fails in our challenging setting, where cameras are sparse and spaced far from each other.  We thus manually specify the reference camera (camera 0, or $C_0$) to lower the challenge.

Formally, given a set of point correspondences $(\bm{p}_i, \bm{p}_0)$ of camera pair ($C_i, C_0$), we first solve the essential matrix $\bm{E}_{i0}$ inside a RANSAC~\cite{fischler1981random} loop.
We then decompose $\bm{E}_{i0}$ into the relative rotation matrix $\bm{R}_{i0}$ and an up-to-scale relative translation $\bm{t}_{i0}$.
Next, we triangulate the 3D points $\bm{P}_{i0}$ using $(\bm{p}_i, \bm{p}_0)$ and \{$\bm{R}_{i0}$, $\bm{t}_{i0}$\}.
After that, we use a local bundle adjustment step to jointly optimize the camera pose \{$\bm{R}^1_{i0}$, $\bm{t}^1_{i0}$\} and 3D points $\bm{P}^1_{i0}$ by minimizing the 2D re-projection error of points from both $C_i$ and $C_0$.  We then use the prior 3D knowledge $\bm{\gamma}$ to solve the scale ambiguity and obtain \{$\bm{R}^2_{i0} = \bm{R}^1_{i0}$, $\bm{t}^2_{i0}$\} and $\bm{P}^2_{i0}$.  The above steps can be mathematically represented as:
\begin{align}
    \bm{E}_{i0} &= \mathrm{essenMatSolver}(\bm{p}_i, \bm{p}_0)\\
    \bm{R}_{i0}, \bm{t}_{i0} &= \mathrm{decomposition}(\bm{E}_{i0})\\
    \bm{P}_{i0} &= \mathrm{triangulation}(\bm{p}_i, \bm{p}_0, \bm{R}_{i0}, \bm{t}_{i0})\\
    \bm{R}^1_{i0}, \bm{t}^1_{i0}, \bm{P}^1_{i0} &= \mathrm{localBA}(\bm{p}_i, \bm{p}_0, \bm{R}_{i0}, \bm{t}_{i0}, \bm{P}_{i0})\\
    \bm{R}^2_{i0}, \bm{t}^2_{i0}, \bm{P}^2_{i0} &= \mathrm{ambiguitySolver}(\bm{R}^1_{i0}, \bm{t}^1_{i0}, \bm{P}^1_{i0}, \bm{\gamma})
\label{eq:geo_pipeline}
\end{align}
After solving the relative poses for each camera pair, we use a global bundle adjustment to further optimize the 3D poses of all cameras.
If a 3D point is only visible from one camera pair, we directly use its 3D coordinates as the input of the global bundle adjustment.
Otherwise, we take the mean of the coordinates solved from different camera pairs and set the mean to be the coordinate of the 3D point. We call this process ``merge 3D points", from which we obtain ${\hat{\bm{P}}}$.
Finally, we optimize the camera poses using the global bundle adjustment initialized from $\{\bm{R}^2_{i0}, \bm{t}^2_{i0}\}$ and ${\hat{\bm{P}}}$:
\begin{align}
    \hat{\bm{P}} &= \mathrm{merge3DPoints}(\{\mathbf{P}^2_{i0}\})\\
    \{\bm{R}^*_{i0}, \bm{t}^*_{i0}\} &= \mathrm{globalBA}(\{\bm{R}^2_{i0}, \bm{t}^2_{i0}\}, \hat{\bm{P}})
\label{eq:global_ba}
\end{align}
in which, $\{\bm{R}^*_{i0}, \bm{t}^*_{i0}\}$ is the final estimated camera pose.

We explain here the 3D information we use to solve the scale ambiguity.
For \textit{\textit{Terrace}} dataset, we use the assumption that the height of a person is 1.75m~\cite{roser2013human}. 
For the \textit{\textit{Basketball}} dataset, we use the length of a standard free throw line (3.6m)~\cite{hartyani2006official}.
For our \textit{\textit{ConstructSite}} dataset, we use the length of a standard construction steel pipe (1.0m).
\begin{table*}[t]
\renewcommand{\arraystretch}{1.2}
\begin{center}
  \begin{tabular}{l|c|c|c|c}
    \toprule
     & \multicolumn{3}{c}{Camera Pose Error (CPE, ($mm, {}^{\circ}$)) $\downarrow$} \\
    \cmidrule(r){2-4}
    \hfil\multirow{-2}{*}{Method} & \multicolumn{1}{c}{\textit{Terrace}} & \multicolumn{1}{c}{\textit{Basketball}} & \multicolumn{1}{c}{\textit{ConstructSite}} & \multirow{-2}{*}{\parbox{2.0cm}{\centering Mean $\downarrow$}}\\
    \midrule
    SIFT~\cite{lowe2004distinctive} + BFM~\cite{jakubovic2018image} & $4599mm, 55.03^{\circ}$ & $6140mm, 65.45^{\circ}$ & $10770mm, 45.66^{\circ}$ & $7170mm, 55.38^{\circ}$\\
    SuperPoint~\cite{detone2018superpoint} + BFM~\cite{jakubovic2018image} & $358mm, 54.68^{\circ}$ & $20065mm, 53.97^{\circ}$ & $2530mm, 50.10^{\circ}$ & $7651mm, 52.92^{\circ}$\\
    WxBS~\cite{mishkin2015wxbs} & $1302mm, 54.14^{\circ}$ & ${8.2e^4mm}, 57.01^{\circ}$ & $7.8e^3mm, 48.91^{\circ}$ & $3.0e^4mm, 53.35^{\circ}$\\
    SuperPoint~\cite{detone2018superpoint} + SuperGlue~\cite{sarlin2020superglue} & $9934mm, 36.96^{\circ}$ & $1.2e^4mm, 29.07^{\circ}$ & $1722mm, 24.08^{\circ}$ & $7885mm, 30.04^{\circ}$\\
    \midrule
    Oracle(Manual-pts) & $390mm, 1.18^{\circ}$ & $\textbf{358mm}, \textbf{0.66}^{\circ}$ & $591mm, 2.08^{\circ}$ & $446mm, 1.31^{\circ}$\\
    \midrule
    Ours (Manual-bbox) & $\textbf{308mm}, \textbf{0.52}^{\circ}$ & ${490mm}, 0.85^{\circ}$ & $\textbf{485mm}, \textbf{1.80}^{\circ}$ & ${427mm}, \textbf{1.06}^{\circ}$\\
    Ours (ReID-bbox) & ${308mm}, {0.52}^{\circ}$ & $410mm, {0.88}^{\circ}$ & ${493mm}, 1.85^{\circ}$ & $\textbf{404mm}, {1.08}^{\circ}$\\
    \bottomrule
  \end{tabular}
\end{center}
\vspace{-5pt}
\caption{\textbf{Camera Pose Error (CPE).} We report the camera position and orientation prediction errors. ``Manual-pts" denotes manually annotated 2d point correspondences, ``Manual-bbox" denotes manually associated bounding box correspondences, ``ReID-bbox" denotes associating bounding box correspondences using re-ID network. (The terms are the same for the following tables unless explicitly stated.) For each dataset, we report the mean position and orientation errors of all cameras.}
\label{tab:cam_pose_err}
\end{table*}
\begin{table*}[t]
\renewcommand{\arraystretch}{1.2}
\begin{center}
  \begin{tabular}{l|c c c|c c c}
    \toprule
     & \multicolumn{3}{c}{Re-Projection Error (RPE, $pixel$) $\downarrow$} & \multicolumn{3}{|c}{Error Resolution Ratio(ERR, $\%$) $\downarrow$} \\
    \cmidrule(r){2-4}\cmidrule(r){5-7}
    \hfil\multirow{-2}{*}{Method} & \multicolumn{1}{c}{\textit{Terrace}} & \multicolumn{1}{c}{\textit{Basketball}} & \multicolumn{1}{c}{\textit{ConstructSite}} & \multicolumn{1}{|c}{\textit{Terrace}} & \multicolumn{1}{c}{\textit{Basketball}} & \multicolumn{1}{c}{\textit{ConstructSite}}\\
    \midrule
    SIFT~\cite{lowe2004distinctive} + BFM~\cite{jakubovic2018image} & 254.40 & 533.15 & 177.90 & $44.17\%$ & $92.56\%$ & $23.41\%$\\
    SuperPoint~\cite{detone2018superpoint} + BFM~\cite{jakubovic2018image} & 53.07 & 9.50 & 130.65 & $9.21\%$ & $1.65\%$ & $17.19\%$\\
    WxBS~\cite{mishkin2015wxbs} & 60.09 & 80.21 & 18381.56 & $10.43\%$ & $13.93\%$ & $2418.63\%$\\
    SuperGlue~\cite{sarlin2020superglue} + SuperPoint~\cite{detone2018superpoint} & 36.57 & 11.03 & 96.38 & $6.35\%$ & $1.91\%$ & $12.68\%$\\
    \midrule
    Oracle(Manual-pts) & \textbf{0.45} & \textbf{0.51} & \textbf{13.45} & $\mathbf{0.08\%}$ & $\textbf{0.09}\%$ & $\textbf{1.78}\%$\\
    \midrule
    Ours (Manual-bbox) & 2.30 & 0.88 & 46.26 & $0.40\%$ & $0.15\%$ & $6.09\%$\\
    Ours (ReID-bbox) & 2.30 & 1.01 & 45.10 & $0.40\%$ & $0.18\%$ & $5.93\%$\\
    \bottomrule
  \end{tabular}
\end{center}
\vspace{-5pt}
\caption{\textbf{Re-projection error (RPE) and error resolution ratio (ERR).}  We report the re-projection error, measured by $pixel$, and error resolution ratio, defined as $\frac{RPE}{\text{min}(H, W)}$, where $(H, W)$ is the video resolution.  Except for the oracle baseline, our method outperforms all the othert baselines that use different hand-crafted/deep features and different matching methods.\vspace{-7pt}}
\label{tab:reproj_err}
\end{table*}
\begin{figure*}
  \centering
  \includegraphics[width=0.95\linewidth]{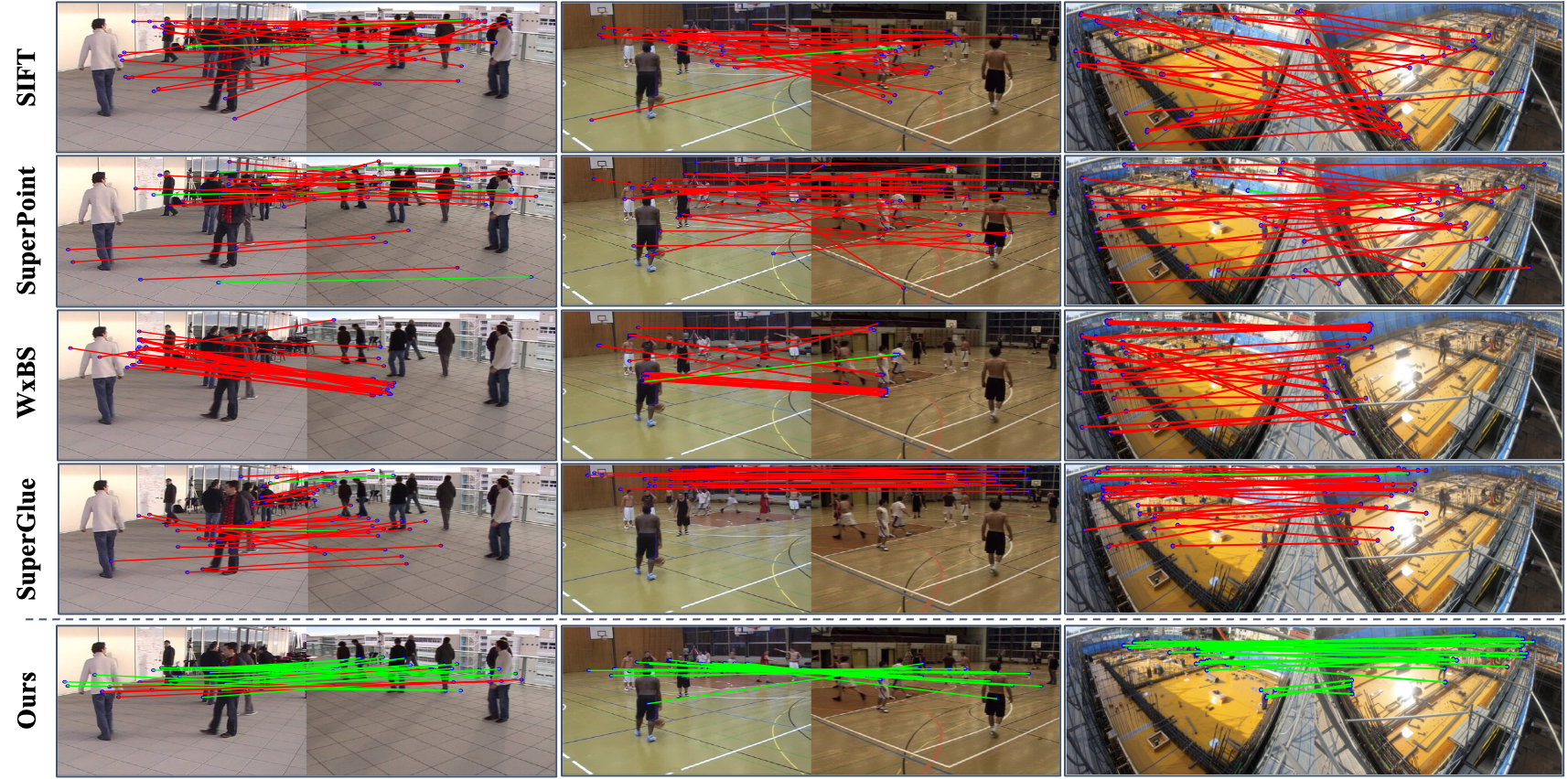}
  \caption{\textbf{Correspondence matching.}  When the distance between a point and the epipolar line computed from its correspondence is less than a threshold (5 pixels), we consider the correspondence as correct and mark the matching as green. The incorrect matchings are marked with red.  Our method performs well under the wide-baseline setting while the baselines fail.\vspace{-10pt}}
  \label{fig:feature_matching}
\end{figure*}
%

\section{Experiment}
We report the evaluation results of our method in this section.
We first describe datasets in Section \ref{subsec:datasets} and evaluation metrics and baselines in Section \ref{subsec:metrics}.
Following in Section \ref{subsec:result}, we present both quantitative and qualitative evaluation results.
Next, we analyze the method robustness and efficiency in Section \ref{subsec:roubust}. Finally, we apply our method in tracking and present the result in Section \ref{subsec:tracking}.

\subsection{Datasets}
\label{subsec:datasets}
We evaluate our method on three datasets captured from scenes of different sizes, camera settings, and human poses.  The camera intrinsic and distortion parameters are provided, and the videos are synchronized for all datasets.

\textbf{{Terrace}}~\cite{fleuret2007multicamera} is an outdoor dataset shot on a terrace outside a building. Up to 7 people evolve in front of 4 DV cameras for 6 minutes 14 seconds. The frame rate is 25 fps, and the video resolution is $720\times576$.  The size of the scene is around $7m\times11m$, and the cameras are about $2m$-high from the ground plane.  People are walking at slow speeds.

\textbf{{Basketball}}~\cite{fleuret2007multicamera} is an indoor dataset filmed at a training session of a local basketball team. It was acquired at a basketball court with 4 DV cameras at 25 fps.  The cameras are about $2m$-high. The videos are 2 minutes and 57 seconds long, with a resolution of $720\times576$.  The size of the scene is about $17.5m\times22m$.  Up to 14 people are doing different activities, including running, jumping, shooting a ball, \etc.

\textbf{{ConstructSite}} is a new dataset collected by our research collaborators using 4 synchronized GoPro HERO7 Black cameras around an outdoor construction site.  The cameras are about $3m$-high from the floor.  The videos are 2 minutes and 57 seconds long, the resolution is $1352\times760$, and the frame rate is 60 fps.  The scene size is about $22m\times28m$.  There are about 20 people in similar suits doing construction works, including standing, crouching, carrying, \etc.

\subsection{Evaluation Metrics and Baselines}
\label{subsec:metrics}

\textbf{Evaluation Metrics}
We use three metrics to measure the performance of our method: (1) Camera pose error (CPE), (2) Re-projection error (RPE), and (3) Error resolution ratio (ERR).  CPE includes the location error and the orientation error.
The location error is the Euclidean distance between the estimated camera location and the ground truth (GT) camera location.
The orientation error is the smallest Euler angle to align the estimated orientation and the GT orientation.
RPE reports the mean re-projection error (by pixel) on 15 pairs of 2D-2D point correspondences that we annotate for each dataset.
ERR reports the ratio between RPE and the video resolution.
ERR (relative) provides a more comprehensive evaluation together with RPE (absolute).

\textbf{Baselines}
We compare our method with the following baselines.
For the first baseline, we detect keypoints from images, use SIFT~\cite{lowe2004distinctive} feature as the descriptor, and match the keypoints across cameras using Brute-Force Matching~\cite{jakubovic2018image} (BFM).  We then use our GeoPose solver to solve the camera pose.
For the second baseline, we use SuperPoint~\cite{detone2018superpoint} network to detect and describe the keypoints simultaneously. We then apply the BFM matching and the GeoPose solver.
The third baseline is WxBS~\cite{mishkin2015wxbs} which uses the idea of view synthesis for wide-baseline matching.  In our setting, ``x" means ``geometry" as defined in the literature.
The fourth baseline is SuperGlue~\cite{sarlin2020superglue}, a method that matches two sets of local features using attention mechanism~\cite{vaswani2017attention} and graph neural networks~\cite{scarselli2008graph}.  We use SuperPoint~\cite{detone2018superpoint} as the feature detector following the paper.
Our last baseline is an oracle baseline.  Specifically, we manually annotate point correspondences and solve camera poses using our GeoPose solver.  We aim to use the oracle baseline to measure the performance gap between our method and standard SfM methods using manual annotations.

\subsection{Comparison with Baselines}
\label{subsec:result}

\textbf{Quantitative results } We present the result of CPE in Table \ref{tab:cam_pose_err} and the result of RPE and ERR in Table \ref{tab:reproj_err}.  
We have the following observations:
(1) All the baselines fail to predict reasonable camera poses on any of the three datasets under our challenging experiment condition.
(2) Our method outperforms all the baselines except for the oracle on all three evaluation metrics.
(3) Our method achieves comparable performance with the oracle on the relatively easier datasets (\textit{\textit{Terrace}} and \textit{\textit{Basketball}}).
(4) Our method performs somewhat worse than the oracle on \textit{\textit{ConstructSite}}, which is a challenging dataset since all the workers wear the same suit, making re-ID mis-associations more likely to happen.
Since the oracle baseline is an SfM pipeline using manual annotations, its performance is the best one can achieve.
Even though our method does not outperform the oracle on \textit{\textit{ConstructSite}}, the result is still encouraging, especially considering the challenging wide-baseline setting and the fact that our method does not require manual annotations.
\begin{table}[t]
\footnotesize
\renewcommand{\arraystretch}{1.3}
\begin{center}
  \begin{tabularx}{\linewidth}{l|l|l|l}
    \toprule
    & \multicolumn{3}{c}{Camera Pose prediction error (CPE) $\downarrow$} \\
    \cmidrule(r){2-4}
    \multirow{-2}{*}{\scriptsize{Noise}} & \multicolumn{1}{c}{\textit{Terrace}} & \multicolumn{1}{c}{\textit{Basketball}} & \multicolumn{1}{c}{\textit{ConstructSite}}\\
    \midrule
    N=0 & $308mm, 0.52^{\circ}$ & ${410mm}, 0.88^{\circ}$ & $493mm, 1.85^{\circ}$\\
    N=1 & $339mm, 0.65^{\circ}$ & $454mm, 1.03^{\circ}$ & $489mm, 1.86^{\circ}$\\
    N=2 & $352mm, 1.00^{\circ}$ & $466mm, 0.95^{\circ}$ & $477mm, 1.83^{\circ}$\\
    N=5 & $407mm, {1.50}^{\circ}$ & ${532mm}, 1.76^{\circ}$ & $537mm, {1.86}^{\circ}$\\\cline{3-3}
    N=10 & ${433mm}, {1.91}^{\circ}$ & ${1653mm, {26.20}^{\circ}}$ & ${601mm}, 1.85^{\circ}$\\\cline{2-2}
    N=20 & ${{1692mm}, {28.49}^{\circ}}$ & $3093mm, {49.20}^{\circ}$ & ${867mm}, 2.31^{\circ}$\\\cline{4-4}
    N=50 & ${4675mm}, {47.80}^{\circ}$ & $3946mm, {64.70}^{\circ}$ & {${{2395mm}, 23.69^{\circ}}$}\\
    \bottomrule
  \end{tabularx}
  \vspace{-10pt}
\end{center}
\caption{\textbf{Noised bounding boxes.} ``N=i" means adding uniformly distributed ($[-i, i]$ pixels) random noise to the top-left and bottom-right corners of the bounding box.  We report camera pose errors with different noise strengths.\vspace{-10pt}}
\label{tab:noise}
\end{table}
%

\textbf{Qualitative results } We present in Figure \ref{fig:feature_matching} the correspondences matching results using our method and the baselines except for the oracle.  We use the distance between a point and the epipolar line (not the re-projection distance) computed from its correspondence to measure the correctness of this correspondence.   If the distance is smaller than a certain threshold, we treat the correspondence as correct. We set the threshold to be 5 pixels and observe that the baselines are not able to correctly match correspondences in wide-baseline scenarios while our method works well.
%
\begin{figure*}
  \centering
  \includegraphics[width=0.95\linewidth]{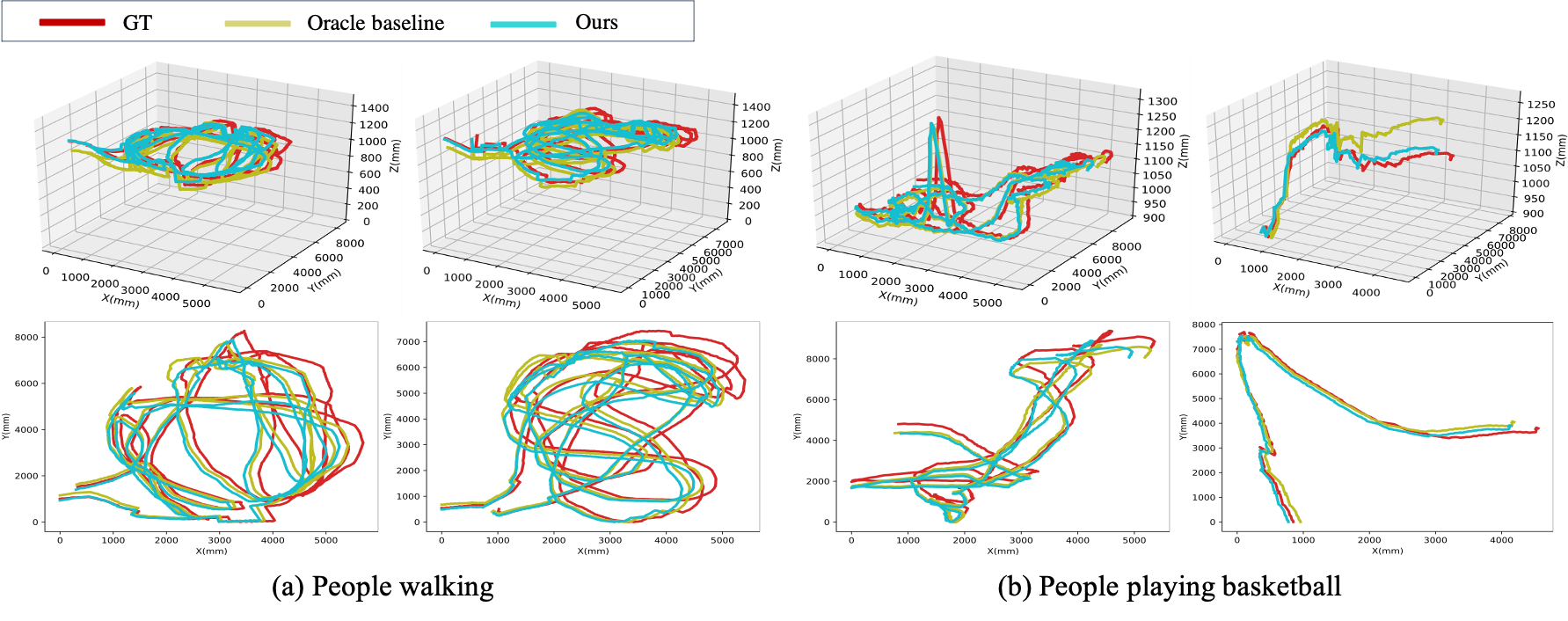}
  \vspace{-0.3cm}
  \caption{\textbf{Tracking results.} We track four people in different motions status (walking, playing basketball) and visualize their 3D trajectories and 2D bird’s-eye view trajectory.  We compare our method with the ground truth (GT) trajectory and the oracle baseline, which solves camera poses using manually annotated 2D-2D point correspondences.  Our method performs similar to the oracle baseline, and the average trajectory difference are $0.28m$, $0.26m$, $0.59m$, and $0.36m$ (from left to right).\vspace{-10pt}}
  \label{fig:tracking}
\end{figure*}
\subsection{Robustness and Efficiency Analysis } 
\label{subsec:roubust}
\textbf{Noised bounding boxes } To evaluate the robustness of our method to the imperfect bounding boxes, we add noises to the person bounding boxes and report camera pose errors on all the three datasets in Table \ref{tab:noise}.  We observe that:  (1) Our method demonstrates a certain level of noise robustness on all the datasets.  (2) For the larger scene (\textit{ConstructSite}) where people are away from the camera, our method shows better noise robustness. We give our understanding here. Imagine that there a ray from the camera to the body mass position in the 3D space.  When the person is away from the camera, a change in the person position will only cause a small direction (or angle) change on the ray. The same amount of position change when the person is close to the camera will lead to a larger change in the ray angle.

\begin{figure}
  \centering
  \includegraphics[width=0.95\linewidth]{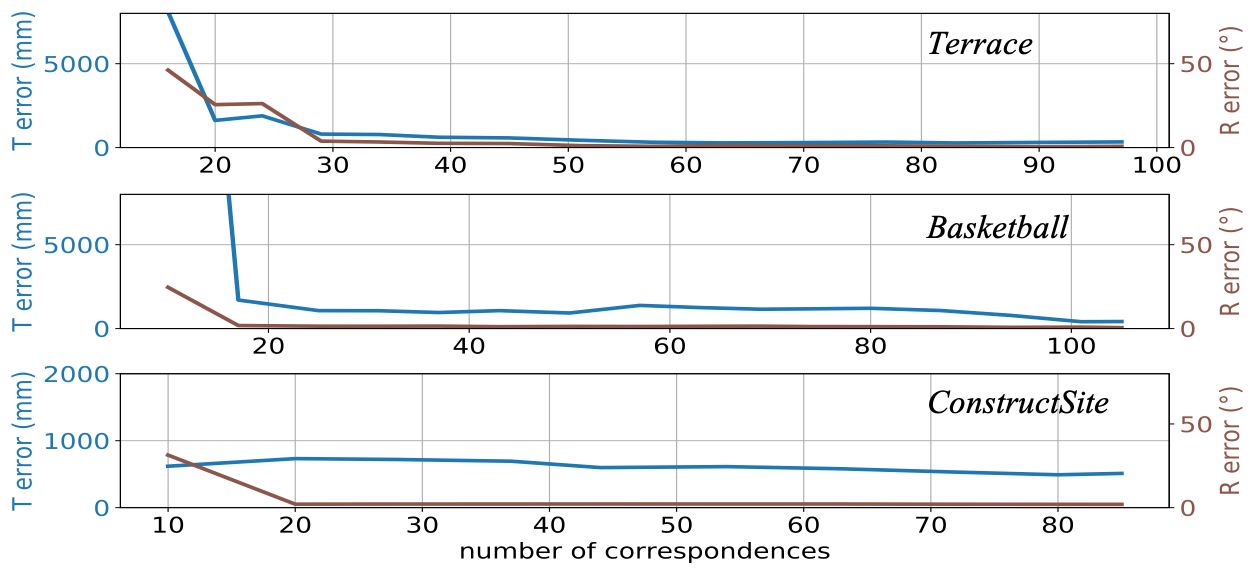}
  \vspace{-5pt}
  \caption{\textbf{CPE \textit{vs} number of correspondences.} We observe that our method converges with the requirement of less than 30 correspondences for all three datasets. \vspace{-20pt}}
  \label{fig:num_corresp}
\end{figure}

\textbf{Number of correspondences } We present in Figure \ref{fig:num_corresp} the plots of CPE $vs$ number of person correspondences used for estimating camera poses.  We observe that our method converges fast as the number of person correspondences increases, and it reaches a good performance with $30$ correspondences for all datasets.
Considering that we use RANSAC and compute the Jacobian matrix in BA in our GeoSolver, using more correspondences is more expensive.  Our method only requires a small number of correspondences to reach good performance.  The cost is low.

\subsection{Application on Tracking}
\label{subsec:tracking}
In this experiment, we apply our camera pose estimation method in the tracking task for estimating the trajectory of a moving person.  We first solve the camera poses using our method.  Next, we specify an object person and estimate the 3D body mass positions of the object person over time using the solved camera poses.  Lastly, we take the mean of the 3D coordinates solved from all camera pairs as the final estimation of the body mass location.  Figure \ref{fig:tracking} shows the trajectories (body mass location over time) of four people from both 3D view and 2D bird's eye view.  We observe that our method consistently gives good trajectory estimation over time (3 mins for walking people, 40s for sporting people).  The difference between the trajectory from our method and the trajectory from the oracle is less than $0.6m$ even for intense body motions like playing basketball.
%

\section{Conclusion}
In this work, we studied the camera pose estimation problem for large-area, wide-baseline camera networks. We contribute a method that treats people as ``keypoints" and applies a re-ID network to obtain 2D-2D  point correspondence for solving camera poses. We evaluated our method on datasets of diverse camera settings and person postures, and our method achieves comparable performance with SfM methods relying on manual annotations. We also provided extensive robustness, efficiency, and applicability analysis. There are still many aspects that need to be explored, such as how to improve robustness towards the imperfectness of the bounding boxes and how to use other objects, \eg, cars, when people are invisible from the scene.

\vspace{2mm}
\noindent\textbf{Acknowledgement: }We thank SHIMIZU CORPORATION for the sponsorship and data creation and Vivek Roy and Zhengyi Luo for their discussion and help on our work.

{\small
\bibliographystyle{ieee_fullname}
\bibliography{egbib}
}

\end{document}